\documentclass[letterpaper, 10 pt, conference]{ieeeconf}  

\IEEEoverridecommandlockouts                              

\usepackage{cite}
\usepackage{amsmath,amssymb,amsfonts}
\usepackage{graphicx}
\usepackage{textcomp}
\usepackage{booktabs}
\usepackage{xcolor}
\usepackage{multirow}
\usepackage{caption, subcaption}
\usepackage{float}
\usepackage{pgfplots}
\usepackage{array}
\usepackage{orcidlink}
\usepackage{booktabs}
\usepackage{graphicx}
\usepackage{array}
\usepackage{siunitx}
\newcommand{\ie}{{\em i.e.}}
\newcommand{\eg}{{\em e.g.}}
\newcommand{\etal}{{\em et al.}}
\newcommand{\etc}{{\em etc.}}

\title{\LARGE \bf
UDA4Inst: Unsupervised Domain Adaptation for Instance Segmentation
}

\author{Yachan Guo$^{1}$, Yi Xiao$^{1}$, Danna Xue$^{1}$, Jose L. Gómez$^{1}$, and Antonio M. López$^{1}$
\thanks{This work has been supported by the Spanish grant Ref. PID2020-115734RB-C21 (ADA/SSL-ADA subproject) funded by MCIN/AEI/10.13039/501100011033. Yachan Guo acknowledges the financial support to her PhD from the Chinese Scholarship Council (CSC), grant number 202208310071. Antonio M. López acknowledges the financial support to his general research activities given by ICREA under the ICREA Academia Program. All authors acknowledge the support of the Generalitat de Catalunya CERCA Program and its ACCIO agency to CVC’s general activities.}
\thanks{$^{1}$All authors are with Computer Vision Center, Universitat Autònoma de Barcelona, Barcelona, Spain.
        {\tt\small yguo@cvc.uab.cat}}%
}

\begin{document}

\maketitle
\thispagestyle{empty}
\pagestyle{empty}

\begin{abstract}
Instance segmentation is crucial for autonomous driving, but is hindered by the lack of annotated real-world data due to expensive labeling costs. Unsupervised Domain Adaptation (UDA) offers a solution by transferring knowledge from labeled synthetic data to unlabeled real-world data. While UDA methods for synthetic to real-world domains (synth-to-real) excel in tasks such as semantic segmentation and object detection, their application to instance segmentation for autonomous driving remains underexplored and often relies on suboptimal baselines. We introduce \textbf{UDA4Inst}, a powerful framework for synth-to-real UDA in instance segmentation. Our framework enhances instance segmentation through \textit{Semantic Category Training} and \textit{Bidirectional Mixing Training}. Semantic Category Training groups semantically related classes for separate training, improving pseudo-label quality and segmentation accuracy. Bidirectional Mixing Training combines instance-wise and patch-wise data mixing, creating coherent composites that enhance generalization across domains. Extensive experiments show UDA4Inst sets a new state-of-the-art on the SYNTHIA~$\rightarrow$~Cityscapes benchmark (mAP 31.3) and introduces results on novel datasets, using UrbanSyn and Synscapes as sources and Cityscapes and KITTI360 as targets. 
Code and models are available at https://github.com/gyc-code/UDA4Inst.

\end{abstract}


\section{Introduction}
Instance segmentation offers detailed, fine-grained object understanding essential for applications such as multi-object tracking, pose estimation, and motion prediction. 
 Specifically, in autonomous driving, it accurately locates safety-critical objects ({\eg}, pedestrians, riders) with precise boundaries, enhancing scene analysis and decision-making
\cite{tseng2021fast,benbarka2023instance,wong2020identifying}. 
However, training such models demands extensive manually labeled data with pixel-level masks, which is labor-intensive and time-consuming, creating a data-hungry bottleneck.

As an alternative, synthetic data offers a solution by providing large-scale, simulator-generated labeled datasets \cite{dosovitskiy2017carla,de2022next}, enabling the creation of infrequent scenarios and rapid annotation. Despite this, models trained on synthetic data often perform poorly in real-world settings due to the domain gap. Unsupervised Domain Adaptation (UDA) methods are crucial for bridging this gap.
While UDA has been widely applied to tasks in image classification \cite{lu2020stochastic}, object detection \cite{garnett2020synthetic,gomez2021co}, and semantic segmentation \cite{hoyer2022daformer,gomez2023co}, its application to instance segmentation in autonomous driving remains limited. Currently, only a few studies such as \cite{zhang2019synthetic,huang2021cross,saha2023edaps} have addressed this, often using suboptimal architectures like Mask R-CNN \cite{he2017mask} with adversarial learning or adapting models designed for panoptic segmentation. 
\begin{figure}
    \centering
    \includegraphics[width=\linewidth]{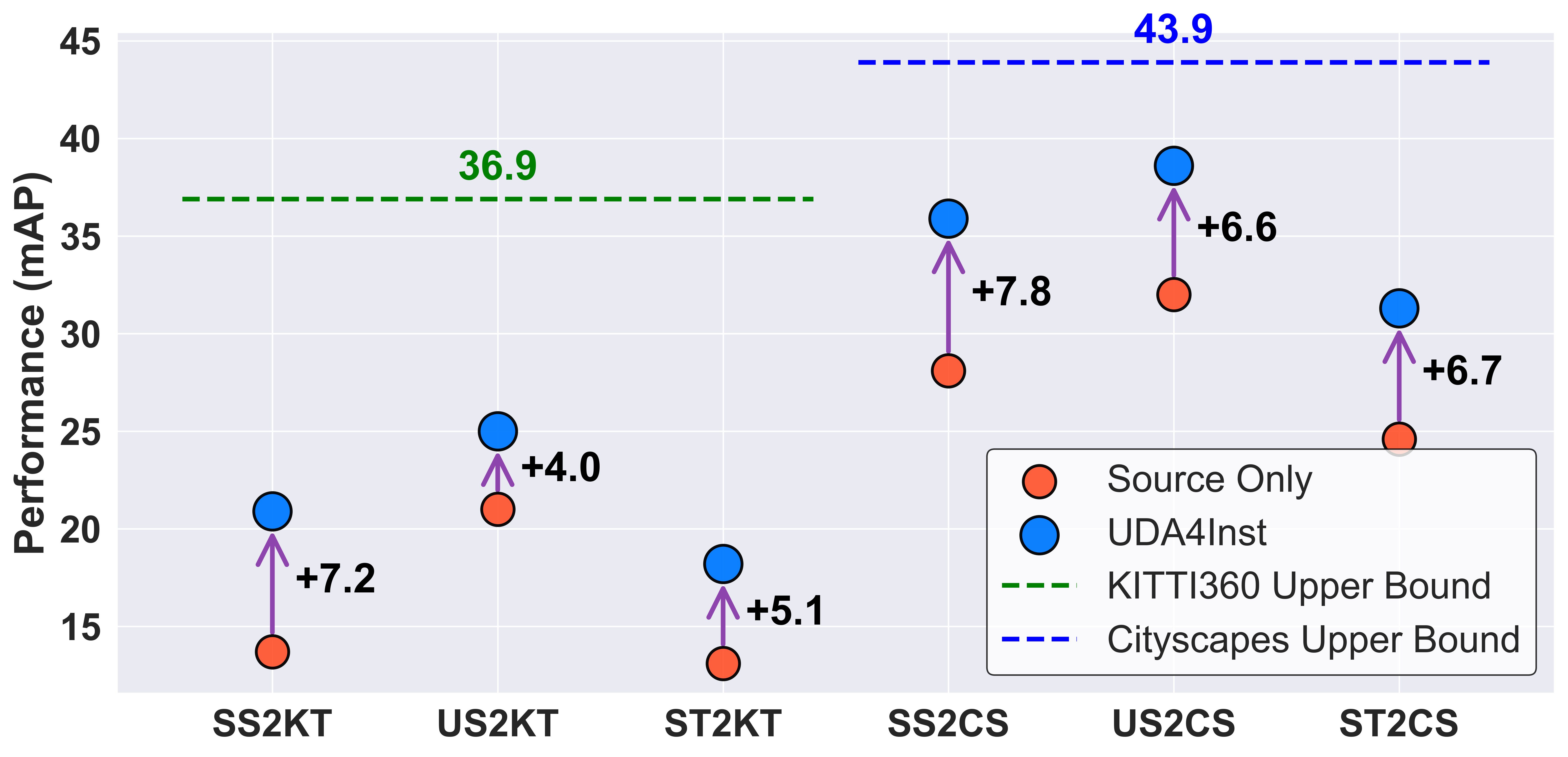}
    \caption{Instance segmentation results for different domain adaptation scenarios from Synscapes (SS), Urbansyn (US), and SYNTHIA (ST) to Cityscapes (CS) and KITTI360 (KT). Dashed lines represent upper-bound models trained on target data, and arrows indicate UDA4Inst's improvements over the source-only baseline.}
    \label{fig:performance_comparison_optimized}
\end{figure}

In this paper, we propose UDA4Inst, a framework for \textbf{UDA for} \textbf{Inst}ance Segmentation, employing a teacher-student online self-training approach that generates trustworthy pseudo-labels from an unlabeled target dataset. 

First, We design a \textit{semantic category training approach}, where we categorize the instance classes by a criterion to train more specialized models. These models use Mask2Former \cite{cheng2022masked} to ensure up-to-date strong instance segmentation baselines. We combine the outputs of these specialized models to obtain better-quality instance masks that will improve the pseudo-labeling generation of our framework. 

Furthermore, we apply a \textit{bidirectional domain mixing training} strategy to
narrow the domain gap between synthetic and real‐world datasets.
Unlike the previous model \cite{Huang_2023_CVPR} that limited the cross-domain data mixing to a single direction, either source-to-target (S2T) or target-to-source (T2S), UDA4Inst combines and balances both strategies to significantly boost instance segmentation performance.

Note that one of the most challenging scenarios in instance segmentation is where multiple instances of the same class overlap. In semantic segmentation, these overlapping instances are treated as a single blob. In contrast, instance segmentation methods need to find clear boundaries in the overlapping region of objects to separate them. We solve this challenge by adopting an instance-wise mixing strategy for large-area instances, ensuring that composite images retain clear object boundaries. For smaller instances, we use a patch-wise mixing method, ensuring that the instances in the mixed images retain the surrounding environmental context. This prevents small, indistinct objects from appearing abruptly or unnaturally in the new domain images.
Additionally, to address instance imbalance, UDA4Inst incorporates a straightforward online algorithm that balances rare-class instances during data mixing between domains.

Finally, as a good practice, we evaluate the quality of the pseudo-labels generated by our system, using them exclusively to train a vanilla model in a supervised fashion on their respective target dataset.

We demonstrate the effectiveness of UDA4Inst on multiple UDA instance segmentation benchmarks. On SYNTHIA~\cite{ros2016synthia}$\rightarrow$Cityscapes,
we achieve state-of-the-art results. To the best of our knowledge, this is the first report performing UDA instance segmentation using the synthetic driving scene datasets UrbanSyn \cite{gomez2025all} and Synscapes \cite{wrenninge2018synscapes}, achieving 38.6 mAP and 35.9 mAP on Cityscapes~\cite{cordts2016cityscapes} and 25.0 mAP and 20.9 mAP on KITTI360~\cite{liao2022kitti}. Compared to the source-only method, our approach demonstrates substantial improvements across these benchmarks as in \textbf{Fig.} \textbf{\ref{fig:performance_comparison_optimized}}. Our contributions are as follows:

\begin{itemize}
\item \textbf{Semantic Category Training}: We propose a novel training strategy for instance segmentation that groups semantically related classes and trains specialized models for each group, yielding better performance than a single model trained on all classes simultaneously.
\item \textbf{Bidirectional Mixing Training}: We develop a bidirectional cross-domain mixing approach, which includes patch-wise and instance-wise strategies to mitigate bias and avoid catastrophic forgetting. By adaptively choosing between instance-wise or patch-wise mixing, handling overlapping masks, aligning color spaces, and rebalancing rare classes, our method effectively tackles the domain gap and enhances instance segmentation performance across synthetic and real-world datasets.
\item \textbf{Raising Performance}: Our UDA4Inst framework establishes a new state-of-the-art with mAP 31.3 on the SYNTHIA$\rightarrow$Cityscapes benchmark, improving by 6.7 points with respect to the baseline and 15.6 points to the previous state-of-the-art method.
\item \textbf{Establishing New Cross-Domain Benchmarks}: We establish additional novel cross-domain settings with UrbanSyn and Synscapes as synthetic source datasets and Cityscapes and KITTI360 as a real-world target, broadening the scope for future research.
\end{itemize}

\section{Related Work}
\subsection{Instance Segmentation}
Early mainstream deep learning models based on convolutional neural networks (CNNs) for instance segmentation, such as Mask R-CNN \cite{he2017mask} and YOLACT \cite{bolya2019yolact}, achieved remarkable successes. 
In recent years, with the adoption of the self-attention mechanism in natural language processing
and computer vision
transformer-based model
\cite{li2023mask} 
have emerged and surpassed CNN-based architectures in many tasks such as object detection, semantic segmentation and panoptic segmentation.
Among them, Mask2Former has attracted great attention as a universal transformer-based architecture that obtains top results in three major image segmentation tasks (panoptic, instance, and semantic) on several benchmarks. It introduces masked attention, which extracts localized features by constraining cross-attention within predicted mask regions. In this work, we leverage this advanced universal segmentation architecture to prompt the state-of-the-art results in instance segmentation with UDA.
\subsection{Unsupervised Domain Adaptation}

Typically, UDA methods fall into three groups: domain discrepancy alignment, adversarial learning, and self-training. In domain discrepancy alignment, a suitable divergence ({\eg}, maximum mean discrepancy \cite{rozantsev2018beyond}, correlation alignment \cite{sun2016return}, or Wasserstein distance \cite{liu2020importance}) is minimized in a latent feature space. Adversarial learning approaches \cite{gong2019dlow} encourage domain invariance at the input, feature, or output level. In contrast, the self-training method \cite{hoyer2023mic} leverages both labeled and unlabeled data, generating high-confidence pseudo-labels for the latter and iteratively refining them to expand the labeled set. 
Our approach follows this self-training paradigm, aligning with mainstream UDA methods and supporting our bidirectional mix strategy. Contemporary UDA research mainly targets semantic segmentation and is nearing fully supervised performance. Conversely, UDA panoptic segmentation is underexplored, with only a few methods \cite{huang2021cross,saha2023edaps} addressing domain-adaptive instance and semantic segmentation in one network.

The most relevant work to ours is \cite{zhang2019synthetic}, which narrows the synthetic-to-real gap for instance segmentation via multi-level feature alignment—global, local, and subtle—using Mask R-CNN with adversarial discriminators. Given recent progress in non-cross-domain instance segmentation and UDA, we aim to improve the UDA instance segmentation baseline by integrating cross-domain data mixing and the advanced, transformer-based Mask2Former network.
\subsection{Cross-domain Data Mixing}
A commonly used data augmentation technique is mixing image content. 
In UDA, this augmentation is usually applied in the form of cross-domain mixing, {\ie} the mixed samples consist of data from both source and target domains. Tranheden {\etal} introduce DACS \cite{tranheden2021dacs}, 
a unidirectional cross-domain data mixing strategy that cuts out semantic classes in the source domain and pastes them into samples in the target domain. Suhyeon {\etal} use a similar method by transferring the tail-class content from source to target to solve the class imbalance problem \cite{lee2021unsupervised}. Gao {\etal} propose a dual soft-paste strategy that not only considers mixing from the source to target images but also within the source domain itself \cite{gao2021dsp}. A similar idea is suggested by Huo {\etal}, while for the dual mixing \cite{huo2023focus}, they focus on the target-to-target domain. Most recently, Kim {\etal} proposed a S2T and T2S cut-and-paste strategy based on predefined image patches \cite{kim2023bidirectional}.
It is noteworthy that these mixing approaches fall into two categories: 

(1) \textit{Class-wise Mixing}. For instance, Cardace {\etal} \cite{cardace2022shallow} only mix semantic classes such as `Car', `Person', and `Pole'. This approach fails to delineate boundaries accurately where multiple instances overlap. 

(2) \textit{Random Patches Mixing.} This approach involves cut-and-paste based on arbitrary patches, resulting in unrealistic images with mixed foreground instances and background areas. 

Unlike the prior works in semantic segmentation, we develop a bidirectional domain mixing strategy for both S2T and T2S mixing instance-wise and patch-wise. In addition, 
to create composite images with more realistic semantic boundaries in the mixing area, we use an instance-wise strategy for mixing larger instances, ensuring clear contours and a natural appearance. For smaller instances, a patch-wise mixing method is applied, allowing them to seamlessly integrate into the surrounding environment without appearing out of place, even in the new domain images. 

\begin{figure}
    \centering
    \includegraphics[width=1\linewidth]{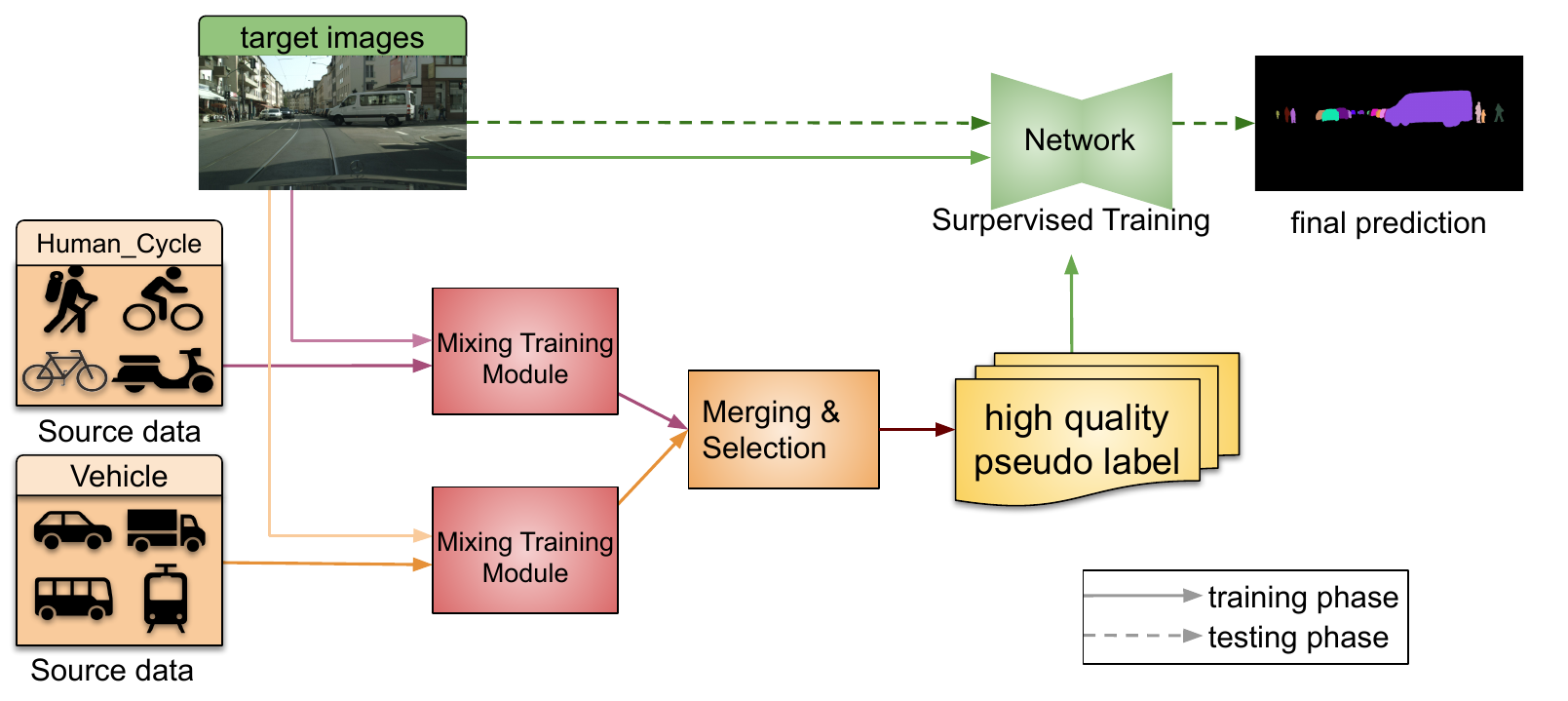}
    \caption{Overview of the UDA4Inst framework. The source data is divided into two categories (Human-Cycle and Vehicle), each independently trained with target images through respective Mixing Training Modules, resulting in two category-specific models. The outputs from these models undergo merging and confidence-based selection to generate high-quality pseudo labels, which are subsequently used for the final supervised training. Solid arrows represent the training phase; dashed arrows denote testing.}
    \label{fig:overview}
\end{figure}

\section{Methodology}
\label{sec:metho}
\subsection{Overall Approach}

Our paper introduces the UDA4Inst framework. \textbf{Fig.} \textbf{\ref{fig:overview}} provides an overview of the pipeline. We perform Semantic Category Training by dividing the training data into two semantic groups, ({\eg} \textbf{Human-Cycle} and \textbf{Vehicle}). We train an individual model for each group in the Mixing Training Module. Moreover, we apply the bidirectional mixing training, which involves cross-domain cut-and-paste operations performed both instance-wise and patch-wise to train the model. The outputs of these models are subsequently merged and selected by confidence-based assessments to generate high-quality pseudo-labels from the unlabeled target data. Our final step uses only these pseudo-labels to train a single final model and then produce the final predictions in the target domain. This framework aims to generate high-quality pseudo-labels, which are used to train with the target domain data in a supervised manner, ultimately producing the final instance segmentation results. After an overall explanation of our UDA4Inst framework, we explain key proposals in more detail: Semantic Category Training and Bidirectional Mixing Training.

\subsection{Semantic Category Training} Instance labels inherently carry distinct semantic meanings. Classes that share similar semantics often exhibit overlapping features. For instance, motorcycles and bicycles display a higher degree of feature similarity compared to trains and bicycles, reflecting closer semantic affinity among certain classes.
We hypothesize that introducing an inductive bias by grouping semantically similar classes can guide the model to more effectively learn both inter-class and intra-class features. Consequently, we propose organizing similar classes into groups and training specialized models for each group to enhance segmentation performance by generating more accurate pseudo-labels in the target domain. In this study, we categorize the eight instance classes into two distinct groups: (1) Human-Cycle, comprising person, rider, bicycle, and motorcycle; and (2) Vehicle, comprising car, truck, bus, and train.

Instead of training all classes at once, our method first groups semantically related classes for specialized training. 
In this framework, we first train separate models, each dedicated to a subset of the class space. This category-based training strategy ensures that each model refines its internal representations for a distinct group of classes. We extend this methodology from synthetic datasets to more complex real-world datasets such as Cityscapes and KITTI360, applying the same principle of focused class grouping.

Building on this strategy, we integrate the category-based approach into an unsupervised domain adaptation (UDA) pipeline. Within the UDA framework, two specialized models are initially trained in parallel, each model responsible for a different set of classes. Their outputs are subsequently fused to produce pseudo-labels for the target domain. The fusion mechanism involves confidence-based assessments of each model’s output, merging segmentation results at the class level, and ensuring that each instance is attributed to one model. By concentrating on distinct class groups, our method promotes robust feature representations and generates higher-quality pseudo-labels for the target domain. Detailed comparisons of full-class and category-driven strategies, as well as ablation studies, are provided in Section \ref{sec:ex-sct}.

\subsection{Bidirectional Mixing Training}
\begin{figure}[tb]
\centering
\includegraphics[width=\linewidth] {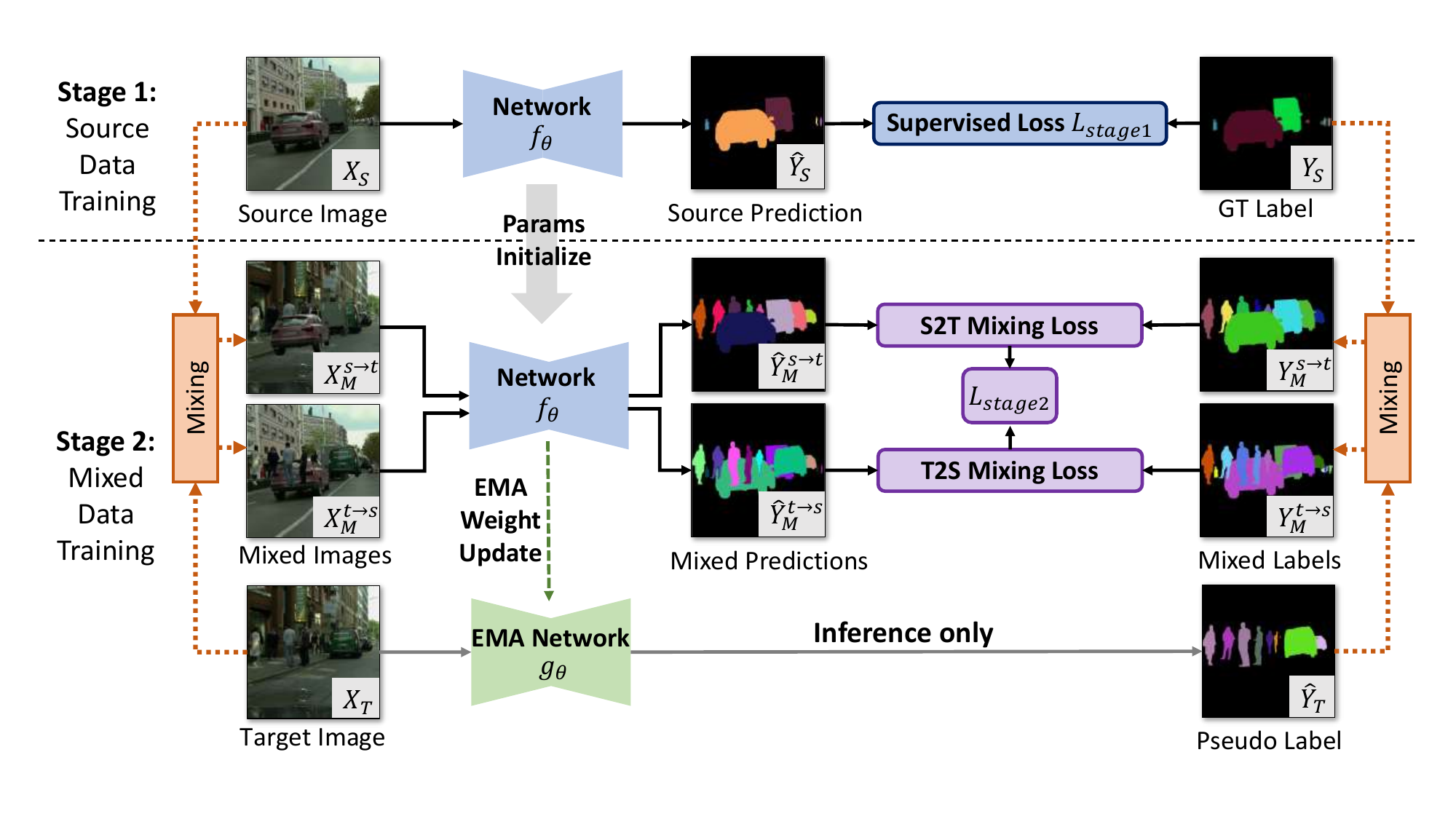}
\caption{Pipeline of mixing training module. In the first stage, the student network is trained with source domain data in a supervised manner. 
In each iteration of the second stage, we update the EMA teacher $g_{\theta}$ with the weights from the network $f_{\theta}$ to predict pseudo-labels in the target domain. Next, the mixed data are generated with the source and target images, and the mixed labels are generated by the target pseudo-labels and the source ground-truth labels. The network is then trained on the mixed data with S2T and T2S mixing losses.
}
\label{fig:mixing_training_module}
\end{figure}

On each of the two categories, we apply our Bidirectional Mixing Training. The training module pipeline is shown in \textbf{Fig.} \textbf{\ref{fig:mixing_training_module}}. For the source domain dataset $\mathcal{D}_S$, comprising images $X_{S}$, their instance labels $Y_{S}$, an instance segmentation model $f_{\theta}$ is trained in a supervised manner. The target domain dataset $\mathcal{D}_T$ only provides images $X_{T}$. We aim to address the domain gap by leveraging $X_{S}$, $Y_{S}$, and $X_{T}$ to train a model capable of accurately segmenting instances on $X_{T}$. 

 Our teacher-student UDA method is based on online self-training. The training has two stages. In the first stage, the model $f$ with parameters $\theta$ is trained on the source domain in a supervised manner for $T_{stage1}$ iterations, with synthetic images $X_{S}$ and their corresponding ground-truth labels $Y_{S}$, to generate predictions $\hat{Y}_{S}$. Our goal is to minimize a supervised segmentation loss
\begin{equation}
\mathcal{L}_{stage1}=\mathcal{L}_{seg}(\hat{Y}_{S},Y_{S}),
\end{equation}
\noindent then the network can be used to predict pseudo-labels, $\hat{Y}_{T}$, on the target domain. In our case, we use the aforementioned Mask2Former as the base network for instance segmentation. The segmentation loss for Mask2Former is

\begin{equation}
\begin{aligned}
\mathcal{L}_{seg}(\hat{Y},Y) =\;& 
\lambda_{ce} \mathcal{L}_{ce}(\hat{Y}, Y) \\
& + \lambda_{bce} \mathcal{L}_{bce}(\hat{Y}, Y) \\
& + \lambda_{dice} \mathcal{L}_{dice}(\hat{Y}, Y)
\end{aligned}
\end{equation}

\noindent where binary cross-entropy loss $\mathcal{L}_{bce}$ and dice loss $\mathcal{L}_{dice}$ are for binary mask segmentation, $\mathcal{L}_{ce}$ is for instance class classification. $\lambda_{bce}$, $\lambda_{dice}$ and $\lambda_{ce}$ are weighting hyper-parameters. 

In the second stage, we utilize the images and labels produced by our bidirectional cross-domain mixed approach on source and target domain data to resume the training of the model $f_{\theta}$ for $T_{stage2}$ iterations. 
Since the target images $X_{T}$ are unlabeled, we utilize the pseudo-labels generated by the exponential moving average (EMA) teacher model $g_{\theta}$. At the beginning of the second stage, the parameters of the EMA teacher are initialized by the weights of $f_{\theta}$. Then, in each training step, we first infer the target images by the EMA teacher to get pseudo-labels $\hat{Y}_{T}$. 
In line with common practice \cite{tranheden2021dacs,hoyer2022daformer,hoyer2023mic}, we use confidence thresholds to filter out noisy pseudo-labels. A pseudo-label confidence score is defined as the product of \textit{mask confidence} (confidence of whether an instance object exists in a mask) and \textit{class confidence} (confidence of an instance class). Then, we leverage the ground-truth labels $Y_{S}$ and the pseudo-labels $\hat{Y}_{T}$ to perform bidirectional instance cut-and-paste, 
thereby obtaining mixed data $X^{s \rightarrow t}_{M}, Y^{s \rightarrow t}_{M}, X^{t \rightarrow s}_{M}, Y^{t \rightarrow s}_{M}$. Finally, mixed data are used to train the model $f_{\theta}$ with a weighted bidirectional data mixing loss
\begin{equation}
    \mathcal{L}_{stage2}=\lambda_{mix}^{s \rightarrow t} \mathcal{L}_{seg}(\hat{Y}^{s\rightarrow t}_{M},Y^{s\rightarrow t}_{M})+ \lambda_{mix}^{t \rightarrow s} \mathcal{L}_{seg}(\hat{Y}^{t\rightarrow s}_{M},Y^{t\rightarrow s}_{M}),
\end{equation}


\noindent where $\lambda_{mix}^{s \rightarrow t}$ and $\lambda_{mix}^{t \rightarrow s}$ are weights for different directional mixing data. Then, the EMA teacher parameters are updated.

\begin{figure}[t]
\centering
\includegraphics[width=\linewidth]{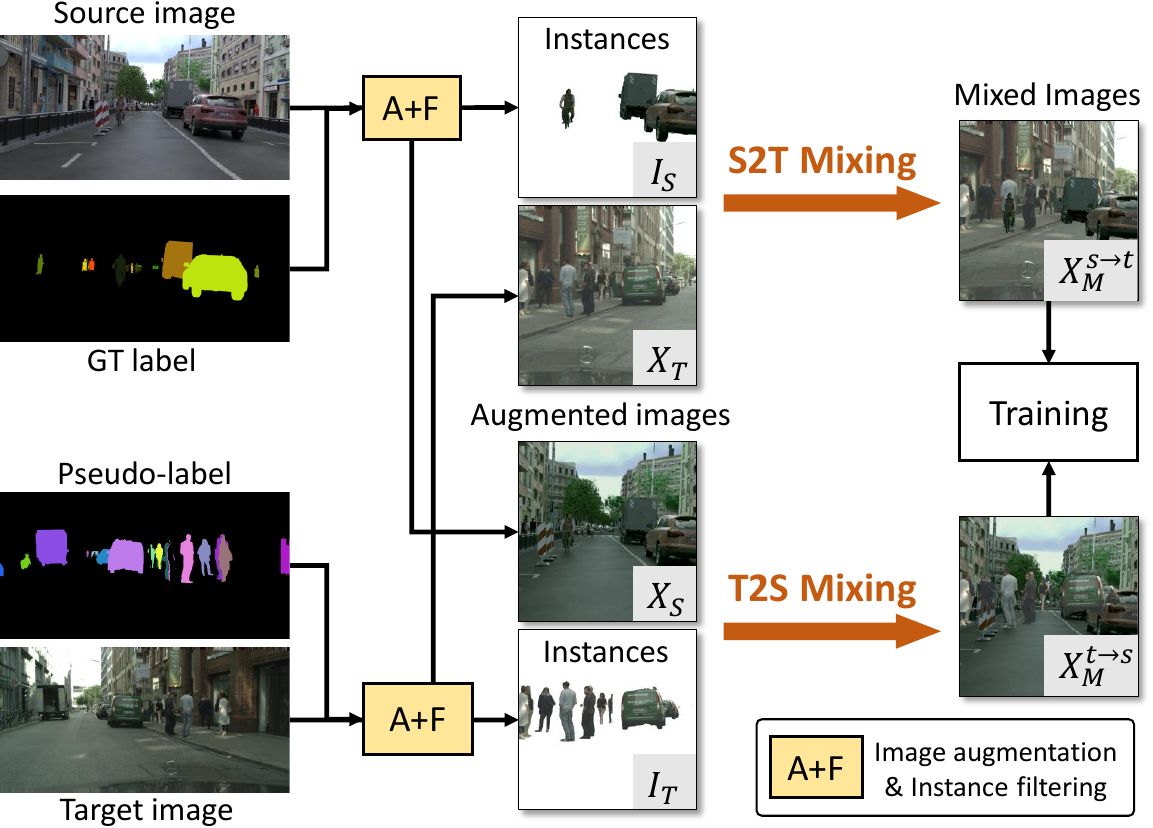}
{\caption{Bidirectional data mixing for instance segmentation. The mixed images and labels are generated by cutting the instances from either the source or target domain, and pasting them to images in another domain.}
\label{fig:datamixing}}
\end{figure}

In data mixing, the mixed images are generated by cut-and-paste image regions from one image to another in a similar fashion to \cite{yun2019cutmix}. We adopt a bidirectional mixing strategy that fully utilizes instances from both source and target domains to facilitate instance segmentation. As shown in \textbf{Fig.} \textbf{\ref{fig:datamixing}}, in each iteration of this training stage, the mixed data is generated in both directions: source to target (T2S) and target to source (T2S). For S2T mixing, the instances in a source domain image $X_{S}$ are cut and pasted onto a target domain image $X_{T}$, resulting in an augmented mixing image $X_{M}^{s \rightarrow t}$. Its corresponding instance label $Y_{M}^{s \rightarrow t}$ is created according to the image, where the instance labels are added to the generated target pseudo-labels $\hat{Y}_{T}$. Similarly, to obtain the T2S mixing data $X_{M}^{t \rightarrow s}$ and $Y_{M}^{t \rightarrow s}$, we leverage the generated pseudo-labels of the target images $\hat{Y}_{T}$ to get the instances that are desired to be pasted on the source image $X_S$. The network is trained on mixed images $X_{M}^{s \rightarrow t}$, $X_{M}^{t \rightarrow s}$ with mixed labels $Y_{M}^{s \rightarrow t}$, $Y_{M}^{t \rightarrow s}$. 

Subsequently, we will elaborate on the strategies adopted during the mixing process.

\subsubsection{Combining Instance-wise and Patch-wise Strategies} 
\begin{figure}
    \centering
    \includegraphics[width=1\linewidth]{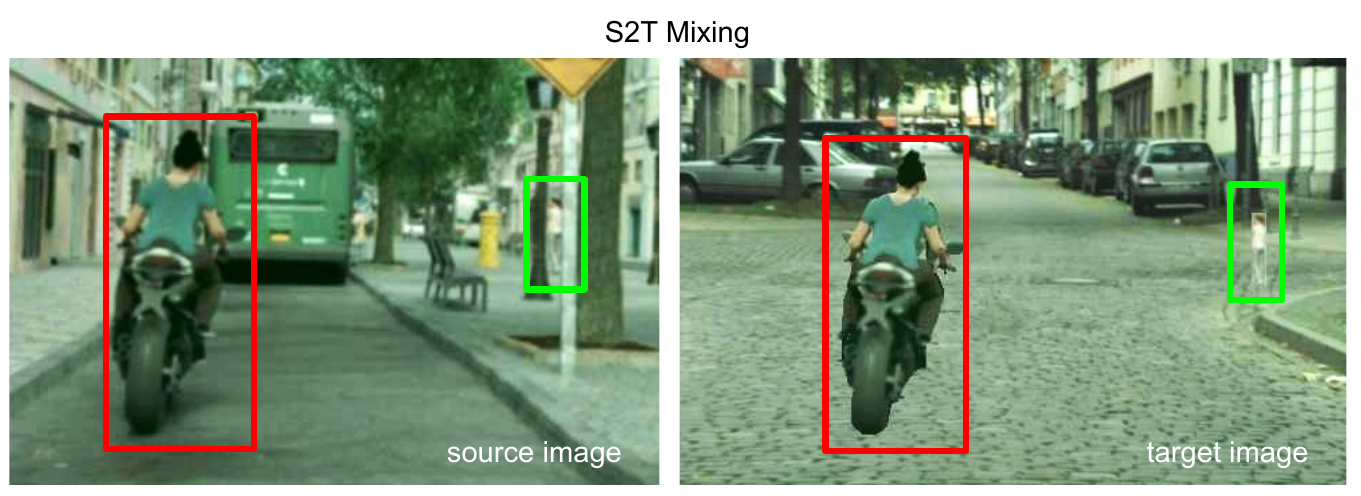}
    \caption{Mixing Strategy. This figure presents the instance-wise and patch-wise combined strategy for different instance sizes from the source image to the target image. \textbf{\textcolor{red}{Red}} bounding box indicates the instance-wise mixing, and \textbf{\textcolor{green}{Green}} one indicates the patch-wise mixing.}
    
    \label{fig:pathch-wise-instance-wise}
\end{figure}

In our bidirectional data mixing strategy, we adaptively choose between instance-wise and patch-wise mixing based on the size of the instances. Specifically, if the area of an instance exceeds a certain threshold, we apply instance-wise mixing. This allows us to paste larger instances onto new backgrounds while preserving clear contours and details, resulting in more coherent composite images. Conversely, for instances whose area is smaller than the threshold, we utilize patch-wise mixing. Small objects often have blurry appearances and unclear boundaries, which makes it difficult to segment them accurately. When such inaccurate small instances are extracted from their original background and pasted into a new domain, they may become difficult to discern due to the lack of contextual information, bringing error to the pseudo-label. In \textbf{Fig.} \textbf{\ref{fig:pathch-wise-instance-wise}}, we show an example of the instance-wise and patch-wise combined strategy for different instance sizes.

\subsubsection{Overlapping Strategy} Overlapping instances in the mixing process can create ambiguous labels and unrealistic visual artifacts, potentially confusing the model during training. To address this, we implement a priority masking strategy. Specifically, in S2T mixing, we assign higher priority to the source domain instances. If an instance from the source image overlaps with an instance in the target image, we erase the overlapping region of the target's instance labels, keeping only the source instance in that region. This means that the source instance's mask replaces any conflicting target instance masks in the overlapping area. We do an analogous strategy for T2S mixing.

\subsubsection{Color Space Transfer} To minimize the color discrepancy, we adopt the same image processing as \cite{he2021multi}, aligning the style of source domain images to the target ones by shifting the distribution of pixel values in CIELAB color space.

\subsubsection{Rare-class Balancing}
Class imbalance is a common issue that affects the performance of deep learning models. Models trained on imbalanced datasets tend to favor classes with more data, leading to lower accuracy in classes with fewer examples.
Therefore, we increase the training frequency of these rare class instances by intentionally resampling them in data mixing.
In each category group, we define the class that occupies the smallest percentage of instances in the source dataset as a rare class. For example, in UrbanSyn, `Motorcycle' and `Train' are 2.47\% and 0.37\%.
Specifically, during the process of traversing images for data mixing, for those images containing rare-class instances, 
we store these images and the corresponding instance labels in a pool with a maximum number of N. These samples are continuously updated with newly emerged rare-class instances and their corresponding images in a FIFO (first-in, first-out) fashion. In turn, when encountering images that do not contain rare classes, we randomly select half of the samples in the pool and mix them on these images. To avoid error accumulation from pseudo-labels, we only apply rare-class balancing to data mixing in the S2T direction.

\section{Experiments}
\label{sec:ex}
\subsection{Datasets}
In this work, we perform experiments across comprehensive synthetic-to-real domain settings. We selected SYNTHIA~\cite{ros2016synthia}, Synscapes~\cite{wrenninge2018synscapes} and UrbanSyn~\cite{gomez2025all} as synthetic datasets and Cityscapes~\cite{cordts2016cityscapes} and KITTI360~\cite{liao2022kitti} as real-world datasets. All the numbers present in the tables are in \%.



\subsection{Implementation Details}
\subsubsection{Training}
We adopt Mask2Former \cite{cheng2022masked} with Swin Transformer backbone \cite{liu2021swin} as our base network $f_{\theta}$. The model is pre-trained on COCO~\cite{lin2014microsoft} dataset.
We apply the same data augmentation (flipping, random cropping, {\etc}) as \cite{cheng2022masked} and set the image crop size to $1024\times1024$ when the target domain is Cityscapes and $384\times672$ for KITTI360. For model stabilization, we employ an exponential moving average (EMA) of parameters with a decay rate 0.999. The maximum number of samples N for the rare-class pool is set to 10, and the pseudo-label confidence threshold $\tau$ is 0.9.
We set the numbers of iterations for source domain training $T_{stage1}$ and mixed data training $T_{stage2}$ to be 40k each. During validation, we follow the same checkpoint selection protocol used in \cite{hoyer2023mic}. For each stage of training, we set a batch size to be 3, considering optimal memory usage. We use AdamW \cite{loshchilov2017decoupled} optimizer with an initial learning rate of 0.0001 and a weight decay of 0.05, as well as the `poly' learning rate schedule with the power of 1.
The hyperparameters of the loss 
$\lambda_{bce}$, $\lambda_{dice}$, and $\lambda_{ce}$ are set to 5.0, 5.0, 2.0 respectively.
During the supervised training phase using pseudo-labels, we adopt the same training configuration. For bidirectional mixing, we set the pixel area threshold to 1500; instances with an area larger than this threshold are processed instance-wise, while those below are handled patch-wise.
For these experiments, UDA4Inst runs on 1 NVIDIA A40 GPU with 48GB of memory.


\begin{table}[tb]
\centering
\scriptsize 
\setlength{\tabcolsep}{1.2pt} 
\caption{Instance segmentation performance.
This table shows UDA4Inst improvements over the vanilla Mask2Former baseline.
`Imp.' denotes improvement.}
\vspace{-2mm}
\resizebox{\columnwidth}{!}{%
\begin{tabular}{c|c|cccccccc|cc}
\toprule
\multicolumn{1}{c|}{Setting} & \multicolumn{1}{c|}{Methods} 
& Person & Rider & Car & Truck & Bus & Train & M.C & B.C & mAP & Imp. \\
\midrule
Cityscapes & Supervised
& 41.1 & 32.4 & 61.4 & 44.8 & 67.5 & 50.7 & 26.8 & 26.6 & 43.9 & - \\
\midrule
\multirow{2}{*}{UrbanSyn$\!\rightarrow\!$CS}
& Source
& 25.1 & 19.5 & 39.8 & 37.6 & 51.5 & 46.7 & 19.2 & 16.2 & 32.0 & - \\
& UDA4Inst
& \textbf{30.4} & \textbf{26.6} & \textbf{48.8} & \textbf{46.3} 
& \textbf{63.0} & \textbf{51.6} & \textbf{22.7} & \textbf{19.4} 
& \textbf{38.6} & \textbf{+6.6} \\
\midrule
\multirow{2}{*}{Synscapes$\!\rightarrow\!$CS}
& Source
& 27.2 & 19.2 & 39.5 & 24.0 & 46.7 & 33.1 & 18.7 & 16.6 & 28.1 & - \\
& UDA4Inst
& \textbf{30.0} & \textbf{26.9} & \textbf{47.7} & \textbf{35.1} 
& \textbf{58.3} & \textbf{47.3} & \textbf{21.9} & \textbf{19.8} 
& \textbf{35.9} & \textbf{+7.8} \\
\midrule
\multirow{2}{*}{SYNTHIA$\!\rightarrow\!$CS}
& Source
& 26.3 & 15.2 & 38.3 & - & 43.0 & - & 13.6 & 10.9 & 24.6 & - \\
& UDA4Inst
& \textbf{28.0} & \textbf{19.9} & \textbf{46.8} & - & \textbf{59.5} & -
& \textbf{18.0} & \textbf{15.8} & \textbf{31.3} & \textbf{+6.7} \\
\midrule
\toprule
KITTI360 & Supervised
& 36.6 & 16.8 & 55.5 & 37.0 & 98.4 & 16.4 & 24.0 & 10.3 & 36.9 & - \\
\midrule
\multirow{2}{*}{UrbanSyn$\!\rightarrow\!$KT}
& Source
& 14.0 & 6.8 & 44.2 & 12.0 & 63.5 & 6.7 & 13.6 & 7.0 & 21.0 & - \\
& UDA4Inst
& \textbf{16.4} & \textbf{7.2} & \textbf{45.4} & \textbf{13.6} 
& \textbf{81.6} & \textbf{13.5} & \textbf{13.9} & \textbf{8.4} 
& \textbf{25.0} & \textbf{+4.0} \\
\midrule
\multirow{2}{*}{Synscapes$\!\rightarrow\!$KT}
& Source
& 15.2 & 5.1 & 44.4 & 6.7 & 17.9 & 0.5 & 14.1 & 6.1 & 13.7 & - \\
& UDA4Inst
& \textbf{20.1} & \textbf{4.8} & \textbf{46.9} & \textbf{13.8} 
& \textbf{60.4} & \textbf{0.2} & \textbf{13.4} & \textbf{7.9} 
& \textbf{20.9} & \textbf{+7.2} \\
\midrule
\multirow{2}{*}{SYNTHIA$\!\rightarrow\!$KT}
& Source
& 3.3 & 2.6 & 32.5 & - & 34.1 & - & 2.8 & 3.5 & 13.1 & - \\
& UDA4Inst
& \textbf{10.3} & \textbf{3.0} & \textbf{41.5} & - & \textbf{47.6} & - 
& \textbf{1.9} & \textbf{4.9} & \textbf{18.2} & \textbf{+5.1} \\
\bottomrule
\end{tabular}
}
\vspace{-2mm}
\label{tab:instance-performance}
\end{table}

\subsection{Experimental Results}

\begin{figure*}[tb]
    \centering
    \includegraphics[width=\linewidth]{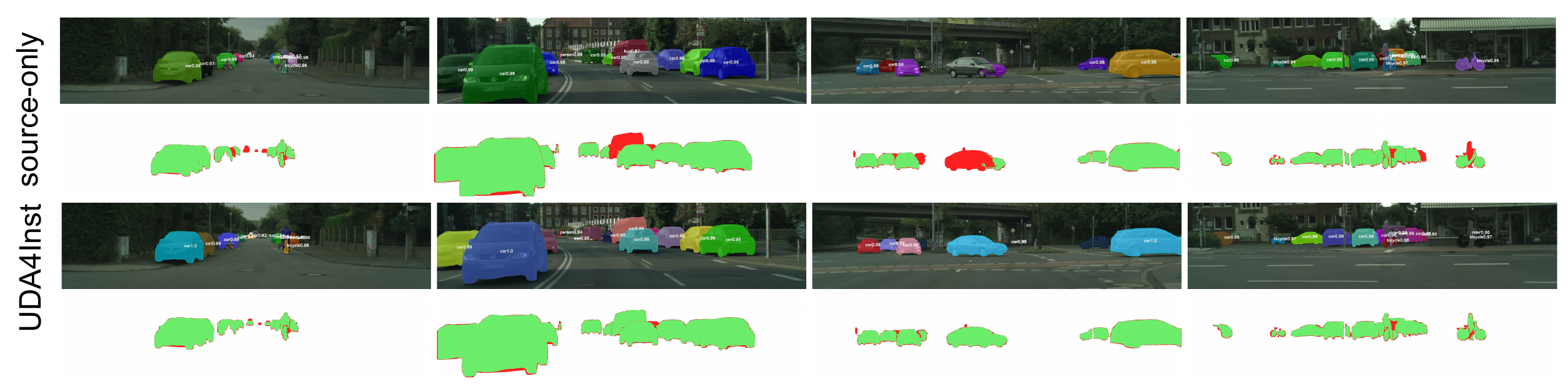}
    \caption{Qualitative comparison of instance segmentation for the source-only model and UDA4Inst for UrbanSyn$\rightarrow$Cityscapes. 
    We show the instance segmentation predictions on images and error maps of the semantic results. \textbf{\textcolor{green}{Green}} indicates correct predictions, while \textbf{\textcolor{red}{red}} highlights errors compared to the ground truth.
    }
    \label{fig:quantitative_uda_vs_baseline}
\end{figure*}

\subsubsection{UDA performance} 

In Table \ref{tab:instance-performance}, we show how UDA4Inst improves over the vanilla Mask2Former baseline in synth-to-real settings. We also include the supervised Mask2Former (trained on Cityscapes and KITTI360) as upper bounds. UDA4Inst consistently outperforms the source-only model, achieving mAP of 38.6 (UrbanSyn$\rightarrow$Cityscapes), 35.9 (Synscapes$\rightarrow$Cityscapes), and a new state-of-the-art 31.3 (SYNTHIA$\rightarrow$Cityscapes)—representing +6.6, +7.8, and +6.7 mAP gains over Mask2Former.
Improvements are particularly notable for larger objects like Car, Truck, Bus, and Train, while smaller, more deformable classes (Person, Motorcycle, and Bicycle) see modest gains. 
Fig. \ref{fig:quantitative_uda_vs_baseline} further illustrates the advantage of UDA4Inst over the source-only model in UrbanSyn$\rightarrow$Cityscapes.
UDA4Inst also shows strong performance on KITTI360, reaching 25.0 mAP (UrbanSyn$\rightarrow$KITTI360) and 20.9 mAP (Synscapes$\rightarrow$KITTI360). Among the source datasets, UrbanSyn consistently performs best on both Cityscapes and KITTI360, likely due to its high photorealism, which narrows the domain gap and boosts adaptation.

\begin{table}[tb]
\setlength{\tabcolsep}{1.4pt} 
\caption{The Instance segmentation performance on SYNTHIA$\rightarrow$Cityscapes.}
\centering
\resizebox{\columnwidth}{!}{%
\begin{tabular}{l|c|cccccccc|c}
\toprule
\multicolumn{1}{c|}{Methods} & \multicolumn{1}{c|}{Backbone} & Person & Rider & Car & Truck & Bus & Train & M.C & B.C & mAP \\ 
\midrule
Zhang~\cite{zhang2019synthetic} 
& ResNet 
& 9.4 & 3.8 & 22.2 & - & 23.5 & - & 3.2 & 2.4 & 10.8 \\


EDAPS~\cite{saha2023edaps} 
& MiT-B5 
& 23.5 & 13.4 & 27.5 & - & 23.6 & - & 5.8 & 0.1 & 15.7 \\ 
\midrule
UDA4Inst 
& Swin-L 
& \textbf{28.0} & \textbf{19.9} & \textbf{46.8} & - & \textbf{59.5} & -
& \textbf{18.0} & \textbf{15.8} & \textbf{31.3} \\
\bottomrule
\end{tabular}%
}
\label{tab:instance_performance_compare_other_paper}
\end{table}

\subsubsection{Comparison with Previous Work} 
UDA for instance segmentation in the autonomous driving field has received limited attention. For reference, in Table \ref{tab:instance_performance_compare_other_paper}, we list results from Zhang \cite{zhang2019synthetic} and EDAPS \cite{saha2023edaps}, which have been proposed in recent years for UDA instance segmentation on the SYNTHIA$\rightarrow$Cityscapes benchmark. 
Zhang \cite{zhang2019synthetic} focuses on instance segmentation using Mask R-CNN with a ResNet backbone, while EDAPS \cite{saha2023edaps} addresses panoptic segmentation, a task that leverages more comprehensive annotations. We present these results to contextualize UDA4Inst’s performance relative to existing approaches, rather than to provide a strictly direct comparison.

\subsection{Ablation Studies}
\subsubsection{Component Ablation}
\label{sec:ex-ablation}
\begin{table}[t]
\caption{Component ablation on UDA methods.}
\setlength{\tabcolsep}{1.2pt}
\centering
\resizebox{\columnwidth}{!}{%
\begin{tabular}{cccc|cccccccc|c}
\\ 
\toprule
 &\multicolumn{1}{l}{SCT} & \multicolumn{1}{l}{BMT} & \multicolumn{1}{l|}{RcB}    & \multicolumn{1}{c}{Person} & \multicolumn{1}{c}{Rider} & \multicolumn{1}{c}{Car} & \multicolumn{1}{c}{Truck} & \multicolumn{1}{c}{Bus} & \multicolumn{1}{c}{Train} & \multicolumn{1}{c}{M.C} & \multicolumn{1}{c|}{B.C} & ~mAP~  \\ 
 \midrule
\multicolumn{1}{c|}{M1}  & - & - & -  & 25.1 & 19.5 & 39.8  & 37.6 & 51.5  & 46.7   & 19.2  & 16.2   & 32.0  \\

\multicolumn{1}{c|}{M2} & \checkmark  & -   & -  & 26.5  & 20.8  & 47.9  & 41.3 & 60.4  & 52.8   & 17.9  & 16.2    & 35.5 \\
\multicolumn{1}{c|}{M3} & -   & \checkmark   & -   & 27.7  & 23.2  & 44.2   & 42.3  & 61.7  & 52.7  & 21.2  & 15.5  & 36.1 \\ 
\multicolumn{1}{c|}{M4} & \checkmark  & \checkmark  & -                        & 28.9  & 27.6  & 50.0  & 43.6   & 63.7   & 47.3   & 19.5  & 20.7   & 37.7  \\
\multicolumn{1}{c|}{M5} & \checkmark & \checkmark  & \checkmark                & 30.4 & 26.6  & 48.8 & 46.3 & 63.0     & 51.6      & 22.7   & 19.4   & 38.6 \\
\bottomrule
\end{tabular}%
}
\label{tab:component-ablation}
\end{table}

Table \ref{tab:component-ablation} shows how UDA4Inst’s three components—Semantic Category Training (SCT), Bidirectional Mixing Training (BMT), and Rare-class Balancing (RcB)—affect UrbanSyn$\rightarrow$Cityscapes.
M1 (no UDA) obtains the lowest mAP of 32.0, while M5 (all components) reaches 38.6.
Comparing M1 vs.\ M2 and M3 vs.\ M4 confirms SCT’s effectiveness, while BMT alone boosts mAP from 32.0 (M1) to 36.1 (M3).
With SCT, BMT further improves M2 by +2.2 mAP.
Adding RcB (M5) yields substantial gains for rare classes like Train and Motorcycle.
Among all components, BMT stands out as the most impactful for adaptation.

\begin{table}[t]
\caption{Ablation study on data mixing strategy.}
\setlength{\tabcolsep}{2pt}
\centering
\begin{tabular}{c|c}
\toprule
\multicolumn{1}{c|}{Mixing Strategy }
& ~mAP~ \\ \midrule
\multicolumn{1}{c|}{w/o mixing} 
& 32.0 \\ 
\multicolumn{1}{c|}{S2T mixing} 
& 35.4 \\ 
\multicolumn{1}{c|}{T2S mixing}
& 35.0 \\ 
\multicolumn{1}{c|}{~bidirectional mixing~}
& \textbf{36.1} \\ 
\bottomrule
\end{tabular}
\label{tab:Data-mixing}
\end{table}

\begin{table}[t]
\caption{Ablation of mixing type. This table shows mAP results of three mixing types, instance-wise, patch-wise, and the combination of both with three source datasets and KITTI360 as the target domain.}
\centering
\setlength{\tabcolsep}{3pt}
\begin{tabular}{c|ccc}
\toprule
Source & instance-wise &  patch-wise &  combined\\ \midrule
UrbanSyn & 14.8 & 21.9 & \textbf{25.0}    \\ 
Synscapes & 15.9 & 12.7  & \textbf{20.9}  \\
SYNTHIA & 12.5 & 13.4   & \textbf{18.2}   \\ 
\bottomrule
\end{tabular}
\label{tab:instance_wise_patch_wise_mixing}
\end{table}

\subsubsection{Bidirectional Mixing Training} We examine data mixing strategies without RcB or SCT in Table~\ref{tab:Data-mixing} (UrbanSyn$\rightarrow$Cityscapes).
Both S2T and T2S mixing help, although T2S is slightly weaker due to noise in target pseudo-labels.
Combining S2T and T2S achieves 36.1 mAP, a +4.1 gain over the baseline without mixing, balancing target learning with pseudo-label noise mitigation.
We also evaluate instance-wise vs.\ patch-wise mixing in Table~\ref{tab:instance_wise_patch_wise_mixing}. For the UrbanSyn, Synscapes, and SYNTHIA source datasets, the combined instance‐wise and patch‐wise mixing strategy outperforms either instance‐wise or patch‐wise mixing alone.

\subsubsection{Impact of Different Class Grouping Strategies} Table~\ref{tab:category} compares four grouped-class models to a full-class model in UrbanSyn$\rightarrow$Cityscapes.
All grouped-class models outperformed the full-class model, with Setting 5 achieving the highest improvement of +3.5 mAP. This validates our hypothesis that semantic similarity-based grouping enhances feature learning and reduces inter-class confusion, thereby improving segmentation accuracy and the quality of pseudo-labels.
Specifically, Setting 5 groups {Person, Rider, Motorcycle, Bicycle} and {Car, Truck, Bus, Train}, providing a balanced and semantically coherent division of classes. In contrast, Setting 2 groups {Person, Rider, Bus, Motorcycle} and {Car, Truck, Train, Bicycle}, which introduces semantic overlap between Bus and Motorcycle. This overlap increases inter-class confusion, resulting in less significant performance gains compared to Setting 5. In the paper, we select setting 5 in our experiments and method explanation. Overall, the ablation study underscores the critical role of semantic similarity in class grouping.

\begin{table}[t]
\caption{Ablation study on Semantic Category Training. The category groups are defined by classes in brackets. }
\setlength{\tabcolsep}{1.2pt}
\centering
\resizebox{\columnwidth}{!}{%
\begin{tabular}{c|c|cc}
\toprule
Setting & Category Groups                                                                   & mAP  & mAP50 \\ \midrule
1       & \{Person, Rider, Motorcycle, Bicycle, Car, Truck, Bus, Train\}              & 32.0   & 54.4  \\
2       & \{Person, Rider, Bus, Motorcycle\}, \{Car, Truck, Train, Bicycle\}  & 33.9 & 57.8  \\
3 & \{Person, Rider, Car, Bicycle\}, \{Truck, Bus, Train, Motorcycle\}  & 35.0 & 59.4  \\
4       & \{Person, Rider\}, \{Car, Truck, Bus, Train, Motorcycle, Bicycle\} & 35.1 & 59.3  \\
5       & \{Person, Rider, Motorcycle, Bicycle\}, \{Car, Truck, Bus, Train\} & 35.5 & 59.4  \\ \bottomrule
\end{tabular}%
}
\label{tab:category}
\end{table}

\subsubsection{Semantic Category Training}
\label{sec:ex-sct}

\begin{table}[t]
  \centering
  \caption{Comparison of full-class vs.\ category-level training (AP only) on Urbansyn, Synscapes, and SYNTHIA (each split into training and validation). “Src” “CS” and “KT” refer to evaluation on the source validation set, Cityscapes, and KITTI360, respectively. The “Imp.” columns show improvements from category-level training. Abbreviations: U-full/U-cat = Urbansyn full/category, S-full/S-cat = Synscapes full/category, SIA-full/SIA-cat = SYNTHIA full/category.}
  \label{tab:full_vs_category}
  \sisetup{detect-all}
  \renewcommand{\arraystretch}{1.1}
  \setlength{\tabcolsep}{1.2pt}
  \newcolumntype{d}[1]{S[table-format=#1]}
  \small
  \resizebox{\linewidth}{!}{
  \begin{tabular}{cc|cccccccc|cc}
    \toprule
    Eval. & Setting
      & Person & Rider & Car & Truck 
      & Bus & Train & M.C & B.C & mAP & Imp. \\
    \midrule
    \multirow{6}{*}{\textbf{Src}} 
    & U-full 
      & 29.9 & 44.1 & 65.9 & 73.0 & 65.1 & {-} & 40.4 & 36.6 & 50.7 & {-} \\
    & U-cat 
      & 48.1 & 46.7 & 71.0 & 79.4 & 68.9 & {-} & 44.0 & 40.6 & 57.0 & +6.3 \\
    & S-full 
      & 47.3 & 50.4 & 59.9 & 57.3 & 57.2 & 50.0 & 43.7 & 40.9 & 50.8 & {-} \\
    & S-cat 
      & 45.6 & 48.0 & 61.7 & 60.6 & 61.0 & 55.4 & 43.0 & 42.4 & 52.2 & +1.4 \\
    & SIA-full 
      & 23.1 & 23.3 & 39.9 & {-} & 40.9 & {-} & 28.5 & 10.5 & 27.7 & {-} \\
    & SIA-cat 
      & 28.3 & 23.7 & 52.1 & {-} & 57.3 & {-} & 29.3 & 10.9 & 33.6 & +5.9 \\
    \midrule
    \multirow{6}{*}{\textbf{CS}} 
    & U-full 
      & 15.4 & 20.3 & 41.7 & 38.9 & 58.0 & 47.1 & 19.2 & 14.5 & 32.0 & {-} \\
    & U-cat 
      & 25.3 & 22.5 & 45.8 & 42.5 & 61.3 & 45.2 & 19.6 & 15.8 & 34.8 & +2.8 \\
    & S-full 
      & 25.9 & 18.3 & 42.9 & 29.2 & 44.5 & 29.5 & 15.1 & 11.9 & 27.2 & {-} \\
    & S-cat 
      & 27.5 & 19.2 & 46.5 & 22.6 & 52.5 & 36.8 & 15.0 & 13.4 & 29.2 & +2.0 \\
    & SIA-full 
      & 9.5 & 16.3 & 37.3 & {-} & 44.0 & {-} & 13.0 & 10.4 & 21.8 & {-} \\
    & SIA-cat 
      & 23.6 & 16.5 & 38.5 & {-} & 44.4 & {-} & 12.4 & 9.2 & 24.1 & +2.3 \\
    \midrule
    \multirow{6}{*}{\textbf{KT}} 
    & U-full 
      & 13.7 & 6.7 & 44.0 & 11.8 & 64.3 & 5.6 & 13.0 & 6.8 & 20.7 & {-} \\
    & U-cat 
      & 15.0 & 6.2 & 44.8 & 10.9 & 79.9 & 12.1 & 12.6 & 6.9 & 23.6 & +2.9 \\
    & S-full 
      & 14.8 & 5.0 & 44.8 & 6.5 & 17.8 & 0.5 & 13.9 & 5.8 & 13.6 & {-} \\
    & S-cat 
      & 13.4 & 4.1 & 42.6 & 5.9 & 45.5 & 0.6 & 12.6 & 5.3 & 16.3 & +2.7 \\
    & SIA-full 
      & 3.0 & 2.1 & 32.3 & {-} & 32.8 & {-} & 2.5 & 2.9 & 12.6 & {-} \\
    & SIA-cat 
      & 4.9 & 3.8 & 35.7 & {-} & 36.5 & {-} & 2.9 & 4.5 & 14.7 & +2.1 \\
    \bottomrule
  \end{tabular}
  }
\end{table}


To verify that grouping semantically related classes produces better pseudo-labels, we compare two Mask2Former-based models: a full-class approach (all eight instance classes trained together) and a category approach (two semantically grouped class sets: Human-Cycle for {Person, Rider, Bicycle, Motorcycle}, and Vehicle for {Car, Truck, Bus, Train}).

We evaluate our method on three synthetic (UrbanSyn, Synscapes, SYNTHIA) and two real domain (Cityscapes, KITTI360) datasets. Each synthetic dataset is split into a training set and a 500-image validation set; Cityscapes also has a 500-image validation set, while KITTI360’s validation set has 12,276 images.

Table \ref{tab:full_vs_category} shows that the category approach consistently surpasses the full-class method across all domains. For instance, it yields a +6.3 mAP boost on UrbanSyn and +5.9 on SYNTHIA. Although improvements on Cityscapes are smaller (up to +2.8 mAP), the category model still generalizes better. Similarly, on KITTI360, the category model outperforms the full-class variant. These results confirm that training specialized models for semantically similar classes enhances both source-domain performance and domain adaptation, making semantic grouping a promising strategy for UDA instance segmentation.

\section{Conclusion}\label{sec:conclusion}
In this paper, we present a teacher-student online framework named UDA4Inst that aims to provide a strong baseline of synth-to-real UDA for instance segmentation. We incorporate effective UDA methods into the vanilla Mask2Former training protocol: Semantic Category Training and Bidirectional Mixing Training. We show that UDA4Inst achieves significant improvements in instance segmentation compared to the source-only Mask2Former in a comprehensive evaluation of synth-to-real domain settings: from UrbanSyn/Synscapes/SYNTHIA to Cityscapes and KITTI360.
Our proposed UDA4Inst model yields state-of-the-art results on SYNTHIA$\rightarrow$Cityscapes. This is the first work to provide competitive average precision (AP) results with Synscapes and the recently published synthetic dataset UrbanSyn.
We expect this work to serve as a reference framework for future research in UDA for instance or panoptic segmentation.

\bibliographystyle{IEEEtran}
\bibliography{egbib}

\end{document}